%% file: main.tex
\documentclass[10pt,twocolumn,letterpaper]{article}

\usepackage{bbding}
\usepackage[accsupp]{axessibility}  

\usepackage{iccv}              

\input{preamble}

\definecolor{iccvblue}{rgb}{0.21,0.49,0.74}
\usepackage[pagebackref,breaklinks,colorlinks,allcolors=iccvblue]{hyperref}

\def\paperID{****} 
\def\confName{ICCV}
\def\confYear{2025}

\title{Momentum-GS: Momentum Gaussian Self-Distillation for High-Quality \\ Large Scene Reconstruction}

\author{
  Jixuan Fan$^{1,2,\ast}$, \quad Wanhua Li$^{3,\ast}$,   \quad Yifei Han$^{1,2}$,   \quad Tianru Dai$^{1,2}$, \quad Yansong Tang$^{1,2,}$\textsuperscript{\Envelope} \\
  $^1$Tsinghua Shenzhen International Graduate School \;
  $^2$Tsinghua University \;
  $^3$Harvard University \\
  \texttt{\small fjx23@mails.tsinghua.edu.cn, wanhua@seas.harvard.edu, hyf23@mails.tsinghua.edu.cn} \\
  \texttt{\small dtr24@mails.tsinghua.edu.cn, tang.yansong@sz.tsinghua.edu.cn} 
}

\begin{document}

\twocolumn[{%
	\renewcommand
	\twocolumn[1][]{#1}%
	\maketitle
	\begin{center}
		\centering
		\vspace{-15pt}
        \includegraphics[width=\textwidth]{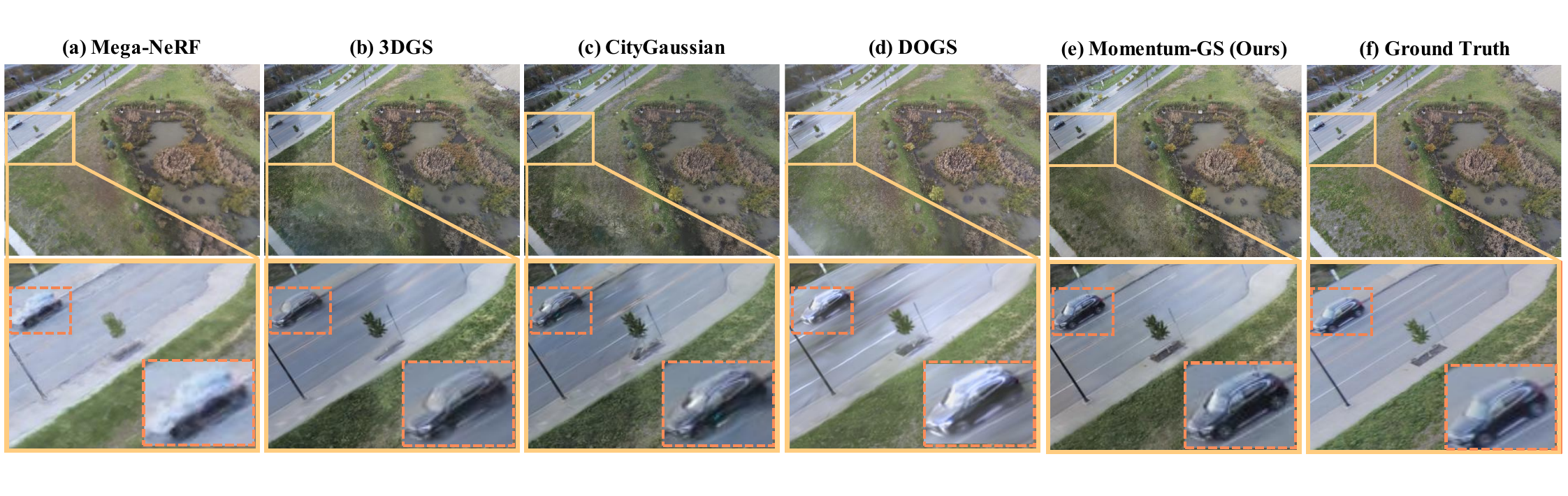}
		\captionof{figure}{\small
             Comparison of reconstruction results on the Rubble \cite{Turki_2022_meganerf} dataset. Our \textbf{\model} reconstructs finer details, such as the clear structure of the vehicle in the zoomed-in view. Additionally, our method produces smoother transitions across blocks, demonstrating better consistency and avoiding the noticeable lighting discrepancies observed in other Gaussian-based methods.}
		\label{fig:teaser}
	\end{center}
}]
{\let\thefootnote\relax\footnotetext{{$^{\ast}$ Equal contribution.\ \textsuperscript{\Envelope}Corresponding authors.}}}

\maketitle

\input{sec/0_abstract}    
\input{sec/1_intro}
\input{sec/2_related_work}

\input{sec/3_methods}
\input{sec/4_experiments}

\input{sec/5_conclusion}

{
    \small
    \bibliographystyle{ieeenat_fullname}
    \bibliography{main}
}

\input{sec/6_suppl}

\end{document}

%% file: preamble.tex
\newcommand{\model}{Momentum-GS\xspace}

%% file: sec/0_abstract.tex
\begin{abstract}
3D Gaussian Splatting has demonstrated notable success in large-scale scene reconstruction, but challenges persist due to high training memory consumption and storage overhead.
Hybrid representations that integrate implicit and explicit features offer a way to mitigate these limitations.
However, when applied in parallelized block-wise training, two critical issues arise since reconstruction accuracy deteriorates due to reduced data diversity when training each block independently, and parallel training restricts the number of divided blocks to the available number of GPUs.
To address these issues, we propose \model, a novel approach that leverages momentum-based self-distillation to promote consistency and accuracy across the blocks while decoupling the number of blocks from the physical GPU count.
Our method maintains a teacher Gaussian decoder updated with momentum, ensuring a stable reference during training. This teacher provides each block with global guidance in a self-distillation manner, promoting spatial consistency in reconstruction.
To further ensure consistency across the blocks, we incorporate block weighting, dynamically adjusting each block’s weight according to its reconstruction accuracy.
Extensive experiments on large-scale scenes show that our method consistently outperforms existing techniques, achieving a 18.7\% improvement in LPIPS over CityGaussian with much fewer divided blocks and establishing a new state of the art. 
Project page: \href{https://jixuan-fan.github.io/Momentum-GS_Page/}{https://jixuan-fan.github.io/Momentum-GS\_Page/}
\end{abstract}

%% file: sec/1_intro.tex
\section{Introduction}
\label{sec:intro}

\begin{figure}[t]
  \centering
   \includegraphics[width=1\linewidth]{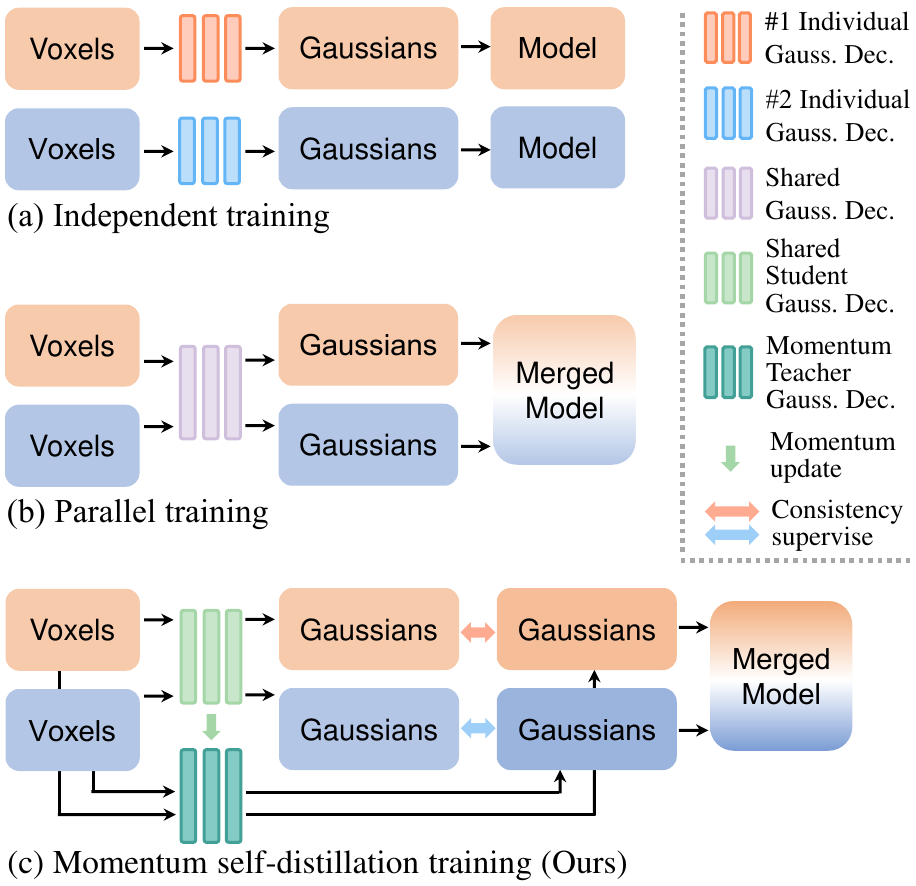}
   \caption{\textbf{Comparison of three approaches for using hybrid representations to reconstruct large-scale scenes in a divide-and-conquer manner.} Examples with two blocks: (a) Independent training of each block, resulting in separate models that cannot be merged due to independent Gaussian Decoders, complicating rendering; (b) Parallel training with a shared Gaussian decoder, allowing merged output but limited by GPU count; (c) Our approach with a Momentum Gaussian Decoder, providing global guidance to each block and improving consistency across blocks.
}
   \label{fig:3solution}
\end{figure}

Large-scale 3D scene reconstruction is essential for a wide range of applications, including autonomous driving~\cite{tancik2022block,li2019aads,ost2021neural,yang2020surfelgan}, virtual reality~\cite{gu2024ue4,jiang2024vr}, environmental monitoring~\cite{yang2024seasplat,liu2024deraings}, and aerial surveying~\cite{bozcan2020air,singh20143d,chen2024survey}. The ability to accurately reconstruct large, complex scenes from collections of images is critical for creating realistic, navigable 3D models and supporting high-quality visualization, analysis, and simulation~\cite{lei2024gauss_nav,chen2024splat_nav,jin2024gs_planner,lu2025manigaussian}.

3D Gaussian Splatting (3D-GS)~\cite{kerbl3Dgaussians} has recently gained attention for its high reconstruction quality and fast rendering speed, outperforming NeRF-based methods~\cite{mildenhall2021nerf,Barron_2021_ICCV,barron2023zip}.
Building on this foundation, recent methods~\cite{lin2024vastgaussian,liu2024citygaussian,yuchen2024DOGS,kerbl2024hierarchical} have further enhanced its performance on large-scale scenes. To handle large environments more efficiently, these approaches often employ a divide-and-conquer strategy that partitions a large scene into multiple independent blocks, allowing for multi-GPU training across these blocks. This method facilitates scalable training for complex, expansive reconstructions.
However, representing millions of Gaussians explicitly creates substantial memory and storage demands~\cite{liu2024citygaussian}, limiting the scalability of 3D-GS for extensive scenes. Additionally, due to unavoidable factors in large scene capture, such as lighting variations, auto-exposure adjustments, or inaccuracies in camera poses~\cite{li2024nerfxl}, independently training each block often disregards inter-block relationships, leading to inconsistencies across block boundaries. This issue can result in visible transitions, as seen in Figure~\ref{fig:teaser} with methods like CityGaussian~\cite{liu2024citygaussian}, where abrupt lighting variations are incorrectly rendered. Addressing these concerns has become a core focus in advancing the field of 3D scene reconstruction.

Hybrid representations~\cite{lu2024scaffold,ren2024octree,li2024ho} have emerged as a promising approach to address memory and storage limitations by combining implicit and explicit features. To manage the complexity of large scenes, these representations integrate dense voxel grids or anchor-based structures with sparse 3D Gaussian fields. These methods typically use MLP as the Gaussian decoder, enabling the generation of neural Gaussians that achieve high reconstruction accuracy while ensuring efficient inference. The decoded Gaussians adapt dynamically to different viewing angles, distances, and scene details.
For instance, in Scaffold-GS~\cite{lu2024scaffold}, during inference, the prediction of neural Gaussians is restricted to anchors within the visible frustum, and trivial Gaussians are filtered out based on opacity using a learned selection process. This approach enables rendering speeds comparable to the original 3D-GS. Additionally, neural Gaussians are generated on-the-fly within the view frustum, allowing each anchor to adaptively predict Gaussians for diverse viewing directions and distances in real time. This adaptive mechanism enhances the robustness of novel view synthesis, delivering high-quality renderings across various perspectives while keeping acceptable computational overhead.

However, applying hybrid representations in parallelized reconstruction for large 3D scenes presents two main challenges. First, training each block independently limits data diversity within each block’s Gaussian decoder, reducing reconstruction quality and producing separate models that cannot be merged due to their independent Gaussian decoders, as illustrated in Figure~\ref{fig:3solution} (a). In contrast, parallel training with a shared Gaussian decoder, as in Figure~\ref{fig:3solution} (b), allows for merging the trained models but constrains scalability, as the number of blocks is limited by the available GPUs. These limitations underscore the need for an approach balancing inter-block consistency and scalability.

To overcome these limitations, we propose \model, a novel approach that combines the benefits of hybrid representations with a strategy tailored to meet the unique demands of large-scale scene reconstruction. Our method decouples the number of blocks from GPU constraints, allowing flexible scaling of reconstruction tasks. This is achieved by periodically sampling \( k \) blocks from a set of \( n \) blocks and distributing them across \( k \) GPUs. To enhance consistency between blocks, we introduce scene momentum self-distillation, where a teacher Gaussian decoder, updated with momentum, provides consistent global guidance to each block, as depicted in Figure~\ref{fig:3solution} (c). This framework encourages collaborative learning across blocks, ensuring that each block benefits from the broader context of the entire scene. Additionally, we introduce reconstruction-guided block weighting, a dynamic mechanism that adjusts the emphasis on each block based on its reconstruction quality. This adaptive weighting enables the shared decoder to prioritize underperforming blocks, enhancing global consistency and preventing convergence to local minima.

To thoroughly evaluate the effectiveness of the proposed method, we conduct extensive experiments on five challenging large-scale scenes~\cite{Turki_2022_meganerf,lin2022urbanscene3d,li2023matrixcity}, including Building, Rubble, Residence, Sci-Art, and MatrixCity. Our \model achieves substantial improvements, demonstrating a 18.7\% gain in LPIPS over CityGaussian~\cite{liu2024citygaussian} while utilizing much fewer divided blocks.

In summary, our contributions are:
\begin{enumerate}
    \item We introduce scene momentum self-distillation to enhances Gaussian decoder performance and decouples the number of divided blocks from the number of GPUs, enabling scalable parallel training.
    \item Our approach incorporates reconstruction-guided block weighting, dynamically adjusting block emphasis based on reconstruction quality to ensure focused improvement on weaker blocks, enhancing overall consistency.
    \item Our approach, \model, achieves better reconstruction quality than state-of-the-art methods, highlighting the strong potential of hybrid representations for large-scale scene reconstruction.
\end{enumerate}

%% file: sec/2_related_work.tex
\section{Related work}
\label{sec:related_work}

\paragraph{Neural Rendering.}
Neural Radiance Fields (NeRF)~\cite{mildenhall2021nerf} have pioneered a breakthrough in novel view synthesis by representing a 3D scene as a continuous volumetric function, where each point along an emitted ray is sampled to produce color and density values. 
Numerous extensions~\cite{Barron_2021_ICCV,pumarola2021dnerf,martin2021nerfinthewild,barron2022mipnerf360,xu2022sinnerf,mildenhall2022nerfinthedark,niemeyer2022regnerf,barron2023zip,reiser2023merf,shen2023sd,tancik2023nerfstudio} have been developed to improve various aspects of NeRF, including its efficiency and scalability.
However, NeRFs require intensive sampling along rays for accurate results, leading to high computational costs and prolonged training and inference times. 3D Gaussian Splatting~\cite{kerbl3Dgaussians} has emerged as a promising alternative, leveraging Gaussian splats for efficient scene representation. Compared to NeRFs, 3DGS significantly reduces sampling requirements while maintaining high fidelity.
It has been widely used for many applications~\cite{qin2024langsplat,chen2024text,li20254d} due to the speed advantage.
Another approach, hybrid representation, combines explicit and implicit elements to benefit from the strengths of both~\cite{muller2022instant,sha2023nerfis,ren2024octree,li2024ho,turki2024hybridnerf}. Often constructed on dense, uniform voxel grids, hybrid representations leverage a mix of methods to improve scene reconstruction. For instance, K-Planes~\cite{fridovich2023k} uses planar factorization to represent multi-dimensional scenes, supporting efficient memory use and applying priors like temporal smoothness. Plenoxels~\cite{fridovich2022plenoxels} adopts a sparse 3D grid with spherical harmonics, bypassing neural networks to directly optimize photorealistic view synthesis from images, achieving significant speedups over traditional radiance fields. Scaffold-GS~\cite{lu2024scaffold} builds on 3D Gaussian Splatting by using anchor points to distribute local 3D Gaussians and predict their attributes dynamically based on viewing direction and distance. These hybrid approaches showcase the advantages of combining explicit and implicit elements for scalable, efficient scene reconstruction.

\paragraph{Large Scene Reconstruction.}
Large-scale scene reconstruction has a long history, with traditional methods often relying on Structure-from-Motion (SfM) ~\cite{sfm,agarwal2011building} to estimate camera poses and create a sparse point cloud from image collections. Subsequent methods, such as Multi-View Stereo (MVS), expanded on this foundation to produce denser reconstructions, advancing the capability of photogrammetry systems to handle large scenes.
With the advent of Neural Radiance Fields (NeRF)~\cite{mildenhall2021nerf}, a shift toward neural representations for photo-realistic view synthesis has enabled more detailed scene reconstructions. Many NeRF-based approaches~\cite{tancik2022block,Turki_2022_meganerf,li2024nerfxl,xu2023grid,zhang2025efficient,mi2023switchnerf}, use a similar divide-and-conquer approach, representing each block independently to facilitate scalable reconstruction. However, these methods still face challenges in rendering speed and consistency across scene segments.
Recently, 3D Gaussian Splatting~\cite{kerbl3Dgaussians} has emerged as a promising alternative, offering real-time rendering with high visual fidelity. 
Numerous methods extend 3DGS to large-scale scenes by enhancing its scalability and efficiency~\cite{wang2024pygs,feng2024flashgs,zhang2024geolrm,ren2024scube,li2024retinags,jiang2024li,cui2024letsgo,liu2024novelgs,chen2024gigagsscalingplanarbased3d,cui2024streetsurfgs}. Some methods~\cite{lin2024vastgaussian,kerbl2024hierarchical,liu2024citygaussian,yuchen2024DOGS,zhang2024garfield++} partition these large scenes into independent blocks for parallel training, allowing for efficient processing and reconstruction .
VastGaussian and CityGaussian, by employing a divide-and-conquer approach to reconstruct large-scale scenes, effectively ensure training convergence, though they lack cross-block interaction, which may limit consistency. DOGS introduces a distributed training method that accelerates 3DGS through scene decomposition and ADMM, while not focusing on optimizing the Gaussian representation for large-scale scenes. These recent 3DGS-based methods demonstrate the potential of 3D Gaussian representations for scalable, high-quality large scene reconstruction, though challenges remain in achieving seamless transitions and efficient memory usage.

%% file: sec/3_methods.tex
\section{Methods}
\label{sec:methods}

\begin{figure*}[ht]
  \centering
   \includegraphics[width=1\linewidth]{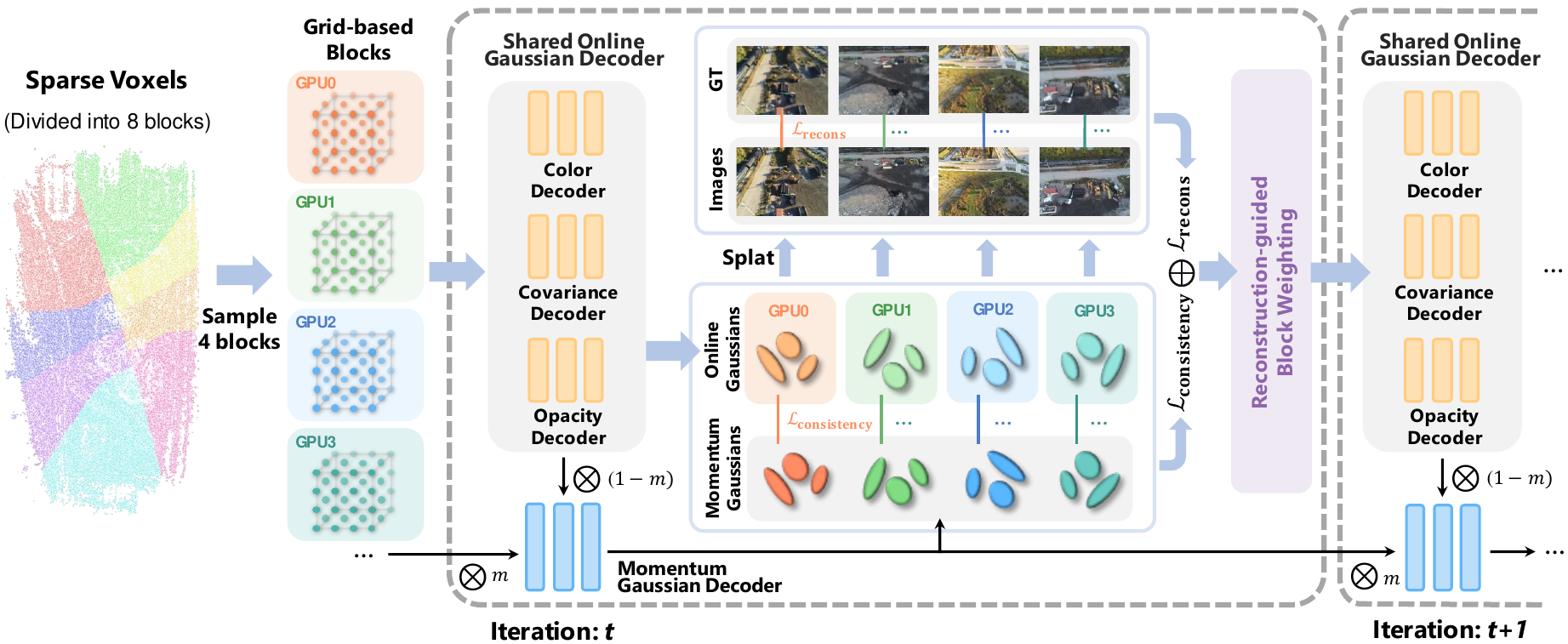}
   \caption{
   \textbf{Overview of the proposed \model.} Our method begins by dividing the scene into multiple blocks (left), periodically sampling a subset of blocks (\eg, 4 blocks) and assigning them to available GPUs for parallel processing. The momentum Gaussian decoder provides stable global guidance to each block, ensuring consistency across blocks. To align the online Gaussians with the momentum Gaussian decoder, a consistency loss is applied. During splatting, predicted images are compared with ground truth images, and the resulting reconstruction loss is used to update the shared online Gaussian decoder. Additionally, reconstruction-guided block weighting dynamically adjusts the emphasis on each block, prioritizing underperforming blocks to enhance overall scene consistency.
}
   \label{fig:pipeline}
   \vspace{-10pt}
\end{figure*}

\paragraph{Overview.}
Hybrid representations have demonstrated success in small, object-centric scenes. However, when applied to parallel training in a divide-and-conquer manner for larger environments, they encounter a fundamental dilemma. In this paper, we leverage hybrid representations for large-scale scene reconstruction, harnessing their high reconstruction capability while effectively decoupling the number of blocks from the physical GPU count.
\Cref{sec:Preliminaries} introduces the essential foundations of 3DGS. \Cref{sec:Scene-Aware_momentum_self-distillation} then explores how Scene Momentum Self-Distillation effectively addresses the challenges of scaling hybrid representations to large scenes. Lastly, \cref{sec:reconstruction-guided_block_weighting} presents the Reconstruction-guided Block Weighting strategy, which enhances global scene consistency by dynamically adjusting each block's weight based on its reconstruction quality.

\subsection{Preliminaries}
\label{sec:Preliminaries}
3DGS offers an efficient solution for accurate scene reconstruction by leveraging the differentiable properties of Gaussian representations along with tile-based rendering. It models each 3D scene point as an anisotropic Gaussian, allowing for streamlined rendering through projection and blending without the computational overhead of dense ray marching typical in traditional volumetric methods.

Each 3D point is represented as a Gaussian function centered at \(\mu \in \mathbb{R}^3\), where \(x\) is the spatial position, \(\mu\) is the center, and \(\Sigma\) defines the Gaussian’s shape and orientation:
\begin{equation}
G(x) = e^{ -\frac{1}{2} (x - \mu)^\top \Sigma^{-1} (x - \mu)}.
\end{equation}
Rendering projects each 3D Gaussian onto the 2D image plane, resulting in a 2D Gaussian \(G'(\mathbf{x}')\), where \(\mathbf{x}'\) represents a pixel. The projected Gaussian contributes to pixel color via alpha blending:
\begin{equation}
C(\mathbf{x}') = \sum_{i \in N} c_i \sigma_i \prod_{j=1}^{i-1} (1 - \sigma_j),
\end{equation}
where \(N\) is the set of Gaussians affecting \(\mathbf{x}'\), \(c_i\) is the color in view-dependent spherical harmonics form, and \(\sigma_i = \alpha_i G'_i(\mathbf{x}')\) is the opacity with \(\alpha_i\) as a learnable parameter.

The training of Gaussians uses differentiable rendering to refine Gaussian parameters, starting from an initial point cloud. Gaussians are optimized based on image reconstruction error, with operations like cloning, densifying, and pruning to improve coverage and accuracy. For large scenes, the high Gaussian count presents memory and computational challenges, managed by controlling the active Gaussians during rendering.

\subsection{Scene-Aware Momentum Self-Distillation}
\label{sec:Scene-Aware_momentum_self-distillation}
Hybrid representations face a fundamental challenge when applied to parallel training in a divide-and-conquer approach. Specifically, the limitation of GPU availability restricts the number of blocks that can be processed simultaneously, reducing scalability, while the need for data diversity to maintain the Gaussian decoder’s predictive accuracy remains critical. To address these challenges, we propose Scene Momentum Self-Distillation, a method that both decouples the block count from GPU limitations and enhances the Gaussian decoder's robustness through improved data diversity. Our method ensures that the Gaussian decoder benefits from a broader range of data, enabling more accurate and consistent predictions across large scenes.

In our approach, we train each block simultaneously in parallel, with all blocks sharing a single Gaussian decoder. During each forward pass, each block randomly selects a viewpoint from its assigned data and uses the shared Gaussian decoder to predict the Gaussian parameters accurately. These predicted parameters are then used to render the corresponding image, which is compared to the ground truth to calculate the reconstruction loss. 
We optimize the learnable parameters using a loss function that combines the $\mathcal{L}_1$ loss on rendered pixel colors with an SSIM~\cite{wang2004ssim} term $\mathcal{L}_{\text{SSIM}}$, aiming to improve structural similarity:
\begin{equation}
    \mathcal{L}_{\text{recons}} = \mathcal{L}_1 + \lambda_{\text{SSIM}} \mathcal{L}_{\text{SSIM}},
\end{equation}

where $\lambda_{\text{SSIM}}$ is a weighting factor that balances the contributions of the $\mathcal{L}_1$ and SSIM terms.
The gradients from each block are accumulated into the shared Gaussian decoder, allowing it to learn from the full range of scene information.

By adopting a sequential training strategy, our method circumvents the GPU-bound constraint on block count. Each GPU handles one block at a time, with periodic switching to ensure coverage of all blocks. This design decouples block quantity from hardware limitations, thereby supporting scalability as scene complexity increases.

To maintain coherence across staggered training blocks and enhance global consistency, we introduce a momentum-based teacher Gaussian decoder \( D_t \) alongside a shared student Gaussian decoder \( D_s \). The Gaussian decoder \( D \) dynamically predicts Gaussian attributes based on viewing positions. Specifically, given the anchor feature \( F \in \mathbb{R}^{N\times32} \), viewing distance \( \delta \in \mathbb{R}^{N\times3} \), and viewing direction \( d \in \mathbb{R}^{N\times1} \), the decoder outputs Gaussian parameters including color, opacity, rotation, and scale. To ensure computational efficiency, the decoder is implemented as a two-layer MLP. Let \( B \) denote the index of each parallel training block, and let \( \theta_t \) and \( \theta_s \) represent the parameters of the teacher and student Gaussian decoders, respectively. We employ a self-supervised approach to stabilize the teacher Gaussian decoder \( D_t \) through momentum-based parameter updates, thus mitigating inconsistencies arising from staggered training. Consequently, the teacher decoder provides a stable global reference, guiding the student decoder via a consistency loss computed between their outputs.

More formally, the parameters \(\theta_t\) of the teacher Gaussian decoder are updated using a momentum-based formula that ensures temporal stability:
\begin{equation}
\theta_t \leftarrow m \cdot \theta_t + (1 - m) \cdot \theta_s,
\end{equation}
where \( m \) is the momentum coefficient, set to 0.9 to balance stability and update speed. If \( m \) is too close to 1, the decoder updates too slowly, hindering reconstruction efficiency, while a smaller \( m \) may lead to instability due to excessive fluctuations in the teacher decoder. 
This momentum-based update ensures that the teacher Gaussian decoder evolves smoothly, providing stable and consistent guidance to the student decoder across all blocks. For each block, Gaussian parameters are predicted by both the teacher and student decoders, with a consistency loss applied to align the student decoder with the global guidance from the teacher. This approach leverages increased data diversity while decoupling the number of blocks from the GPU count, allowing scalability to arbitrarily large scenes.

The consistency loss is computed as the mean squared error between the predictions of the teacher and student Gaussian decoders for each block \( B \):
\begin{equation}
\mathcal{L}_{\text{consistency}} = \| D_m(f_b, v_b; \theta_t) - D_o(f_b, v_b; \theta_s) \|_{2},
\end{equation}
where \( f_b \) represents the anchor feature and \( v_b \) the relative viewing direction for each sample within block \( B \). This loss encourages the student decoder \( D_o \) to progressively align with the stable global guidance provided by the teacher decoder \( D_m \), promoting spatial consistency across different blocks throughout the reconstruction process.

Thus, the total loss function is defined as:
\begin{equation}
\mathcal{L} = \mathcal{L}_1 + \lambda_{\text{SSIM}} \mathcal{L}_{\text{SSIM}} + \lambda_{\text{consistency}} \mathcal{L}_{\text{consistency}},
\end{equation}
where \(\lambda_{\text{consistency}}\) is a weighting factor that balances the impact of the consistency loss relative to the reconstruction loss. This combined loss ensures that the model not only reconstructs the scene accurately but also maintains global spatial coherence across blocks.

\subsection{Reconstruction-guided Block Weighting}
\label{sec:reconstruction-guided_block_weighting}
In order to balance training progress across blocks and mitigate issues arising from uneven initial scene partitioning, we introduce Reconstruction-guided Block Weighting. This method dynamically adjusts weights based on each block’s reconstruction quality, enhancing consistency by giving priority to blocks with lower reconstruction accuracy.

To monitor and adjust the reconstruction performance of each block, we maintain a table that tracks key reconstruction metrics, specifically PSNR (Peak Signal-to-Noise Ratio) and SSIM (Structural Similarity Index). These metrics provide quantitative measures of reconstruction quality, with higher values indicating better visual fidelity. Block PSNR is defined as the average PSNR of every image within a block. Block SSIM is calculated similarly. To ensure that these metrics reflect stable performance across training iterations, we update them using a momentum-based approach, which smooths fluctuations and provides a more reliable indication of each block's progress.

Using these momentum-smoothed metrics, we identify the block with the highest reconstruction performance, labeling its PSNR and SSIM values as \(\mathrm{PSNR_{max}}\) and \(\mathrm{SSIM_{max}}\), respectively. These reference values serve as benchmarks for evaluating the relative accuracy of each block. For every block in the scene, we calculate deviations \(\delta_p\) and \(\delta_s\) to quantify how closely its reconstruction aligns with the highest-performing block. Specifically, the PSNR deviation \(\delta_p\) is obtained by subtracting the current block’s PSNR from \(\mathrm{PSNR_{max}}\). \(\delta_s\) is derived similarly. 

With these deviations calculated, we assign each block a weight \(w_i\) that reflects its relative reconstruction performance. 
The weight \( w_i \) is constructed to resemble a Gaussian distribution, placing greater emphasis on blocks with larger deviations from the best-performing block. By prioritizing blocks with lower reconstruction accuracy, this approach directs the model's attention to underperforming blocks, helping to improve overall consistency across the scene. Additionally, \( w_i \) is capped within a range slightly above one, which prevents excessively high adjustments, ensuring stable training dynamics and avoiding over-penalization of blocks with moderate deviations.

\vspace{-5pt}
\begin{align}
\label{eq:block_weighting}
w_i &= 2 - \exp\left(\frac{\delta_{p}^2 + \lambda \cdot \delta_{s}^2}{-2\sigma^2}\right),
\end{align}
\vspace{-5pt}

This design guiding the Gaussian decoder to focus on the global scene rather than converging on blocks with locally high-quality reconstructions. Consequently improves consistency across all blocks, ultimately enhancing the overall scene reconstruction quality.

%% file: sec/4_experiments.tex
\begin{table*}[ht!]
	\begin{center}
		\resizebox{\linewidth}{!}{
			\begin{tabular}{l|ccc|ccc|ccc|ccc|ccc}
				\toprule
				Scene   &   \multicolumn{3}{c|}{\emph{Building}}  &   \multicolumn{3}{c|}{\emph{Rubble}} &   \multicolumn{3}{c|}{\emph{Campus}}  &   \multicolumn{3}{c|}{\emph{Residence}} &  \multicolumn{3}{c}{\emph{Sci-Art}} \\
				\midrule
				Metrics &  PSNR \(\uparrow\) & SSIM \(\uparrow\) & LPIPS \(\downarrow\) &  
                PSNR \(\uparrow\) & SSIM \(\uparrow\) & LPIPS \(\downarrow\) &
				PSNR \(\uparrow\) & SSIM \(\uparrow\) & LPIPS \(\downarrow\) &
				PSNR \(\uparrow\) & SSIM \(\uparrow\) & LPIPS \(\downarrow\) &
				PSNR \(\uparrow\) & SSIM \(\uparrow\) & LPIPS \(\downarrow\)  \\
				\midrule
				Mega-NeRF~\cite{Turki_2022_meganerf}              & 20.93 & 0.547 & 0.504     & 24.06  & 0.553 & 0.516     & 23.42 & 0.537 & 0.636     & 22.08 & 0.628 & 0.489     & \underline{25.60} & 0.770 & 0.390   \\
				Switch-NeRF~\cite{mi2023switchnerf}               & 21.54 & 0.579 & 0.474     & 24.31 & 0.562  & 0.496    & 23.62 & 0.541 & 0.616     & 22.57 & 0.654 & 0.457     & \textbf{26.52} & 0.795 & 0.360   \\
				3D-GS~\cite{kerbl3Dgaussians}                     & 22.53 & 0.738 & 0.214     & 25.51 & 0.725 & 0.316     & 23.67 & 0.688 & 0.347     & 22.36 & 0.745 & 0.247     & 24.13 & 0.791 & 0.262   \\
				$\text{VastGaussian}$~\cite{lin2024vastgaussian}  & 21.80 & 0.728 & 0.225     & 25.20 & 0.742 & 0.264     & 23.82 & 0.695 & 0.329     & 21.01 & 0.699 & 0.261     & 22.64 & 0.761 & 0.261    \\
				CityGaussian~\cite{liu2024citygaussian}           & 22.70 & \underline{0.774} & 0.246     & \underline{26.45} & \underline{0.809} & \underline{0.232}      & 22.80 & 0.662 & 0.437     & \underline{23.35} & \underline{0.822} & \underline{0.211}     & 24.49 & \underline{0.843} & 0.232    \\
				DOGS~\cite{yuchen2024DOGS}                        & 22.73 & 0.759 & \underline{0.204}     & 25.78 & 0.765 & 0.257     & \underline{24.01} & 0.681 & 0.377     & 21.94 & 0.740 & 0.244     & 24.42 & 0.804 & \underline{0.219}     \\
				\midrule
				\textbf{\model~(Ours)}                            & \textbf{23.65} & \textbf{0.813} & \textbf{0.194}      & \textbf{26.66} & \textbf{0.826} & \textbf{0.200}     & \textbf{24.34} & \textbf{0.760} & \textbf{0.290}     & \textbf{23.37} & \textbf{0.828} & \textbf{0.196}     & 25.06 & \textbf{0.860} & \textbf{0.204}    \\
				\bottomrule
			\end{tabular}		
		}
		\caption{Quantitative comparison of our \model against prior methods across four large-scale scenes. We present metrics for PSNR$\uparrow$, SSIM$\uparrow$, and LPIPS$\downarrow$ on test views. The \textbf{best} and \underline{second best} scores are highlighted.}
		\label{tab:main_table1}
	\end{center}
	\centering
    \vspace{-20pt}
\end{table*}

\begin{table}[t!]
	\begin{center}
		\resizebox{0.8\linewidth}{!}{
			\begin{tabular}{l|ccc}
				\toprule
				Method & PSNR \(\uparrow\) & SSIM \(\uparrow\) & LPIPS \(\downarrow\) \\
				\midrule
				3D-GS~\cite{kerbl3Dgaussians}             & 27.36 & 0.818 & 0.237  \\
				VastGaussian~\cite{yuchen2024DOGS}        & 28.33 & 0.835 & 0.220  \\
				CityGaussian~\cite{liu2024citygaussian}   & 28.61 & 0.868 & 0.205  \\
				DOGS~\cite{yuchen2024DOGS}                & 28.58 & 0.847 & 0.219  \\
				\midrule
				\textbf{\model~(Ours)}                  & \textbf{29.11} & \textbf{0.881} & \textbf{0.180} \\
				\bottomrule
			\end{tabular}		
		}
		\caption{Quantitative comparison on the extremely large-scale urban scene, \textit{MatrixCity}. We report PSNR$\uparrow$, SSIM$\uparrow$, and LPIPS$\downarrow$ on test views, with the \textbf{best} results highlighted.}
		\label{tab:main_table2}
	\end{center}
	\centering
    \vspace{-30pt}
\end{table}

\begin{table}[t!]
	\begin{center}
		\resizebox{\linewidth}{!}{
			\begin{tabular}{l|ccccc}
				\toprule
				Method & 3D-GS & VastGaussian & CityGaussian & DOGS & \textbf{\model~(Ours)}  \\
				\midrule
				FPS \(\uparrow\) & 45.57 & 40.04 & 26.10 & 48.34 & \textbf{59.91}  \\
                \midrule
				Mem \(\downarrow\) & 6.31 & 6.99 & 14.68 & 5.82 & \textbf{4.62}  \\
				\bottomrule
			\end{tabular}		
		}
		\caption{We report the allocated memory (in GB) and rendering framerate (in FPS) measured during evaluation on the extremely large scene \textit{MatrixCity}, with the \textbf{best} results highlighted.}
		\label{tab:FPS}
	\end{center}
	\centering
    \vspace{-20pt}
\end{table}

\section{Experiments}
\label{sec:experiments}

\subsection{Experimental Setup}
\paragraph{Dataset and Metrics.}
We conducted experiments on six large-scale scenes across three aerial drone-captured datasets: \textit{Building} and \textit{Rubble} from the Mill19 dataset~\cite{Turki_2022_meganerf}, \textit{Campus}, \textit{Residence}, and \textit{Sci-Art} from the UrbanScene3D dataset~\cite{lin2022urbanscene3d}, and Small City from the \textit{MatrixCity} dataset~\cite{li2023matrixcity}. Each dataset contains thousands of high-resolution images, with the \textit{MatrixCity} scene notably covering an extensive area of 2.7 square kilometers. Following previous methods~\cite{Turki_2022_meganerf, lin2024vastgaussian, liu2024citygaussian, yuchen2024DOGS}, we downsampled both training and test images by a factor of 4 for all scenes except \textit{MatrixCity}, for which we resized the image width to 1,600 pixels.
We evaluated reconstruction accuracy using PSNR, SSIM~\cite{wang2004ssim}, and LPIPS~\cite{zhang2018lpips}, and additionally reported allocated memory and rendering framerate during evaluation to compare rendering performance.

\vspace{-10pt}

\paragraph{Implementations and Compared methods}
We use the sparse point cloud from COLMAP~\cite{schonberger2016sfm} as our initial input. Each sparse voxel is initialized using a corresponding point from the point cloud.
Following previous methods~\cite{lin2024vastgaussian, liu2024citygaussian}, each block was optimized for 60,000 iterations. To ensure fair comparisons, we use the same initial point cloud and scene partitioning strategy as CityGaussian, but with significantly fewer blocks. Specifically, we divided all scenes into 8 blocks.
Additionally, we applied the same color correction method as DOGS~\cite{yuchen2024DOGS} when computing metrics.
For CityGaussian, we used the checkpoints released by the authors and applied the same color correction method for evaluation. 
We compared our method against Mega-NeRF~\cite{Turki_2022_meganerf}, Switch-NeRF~\cite{mi2023switchnerf}, 3D-GS~\cite{kerbl3Dgaussians}, VastGaussian~\cite{lin2024vastgaussian}, CityGaussian~\cite{liu2024citygaussian}, and DOGS~\cite{yuchen2024DOGS}. All experiments are conducted on the Nvidia RTX 3090 GPUs with 24 GB memory.

\subsection{Results Analysis}

\begin{figure*}[ht]
  \centering
   \includegraphics[width=1\linewidth]{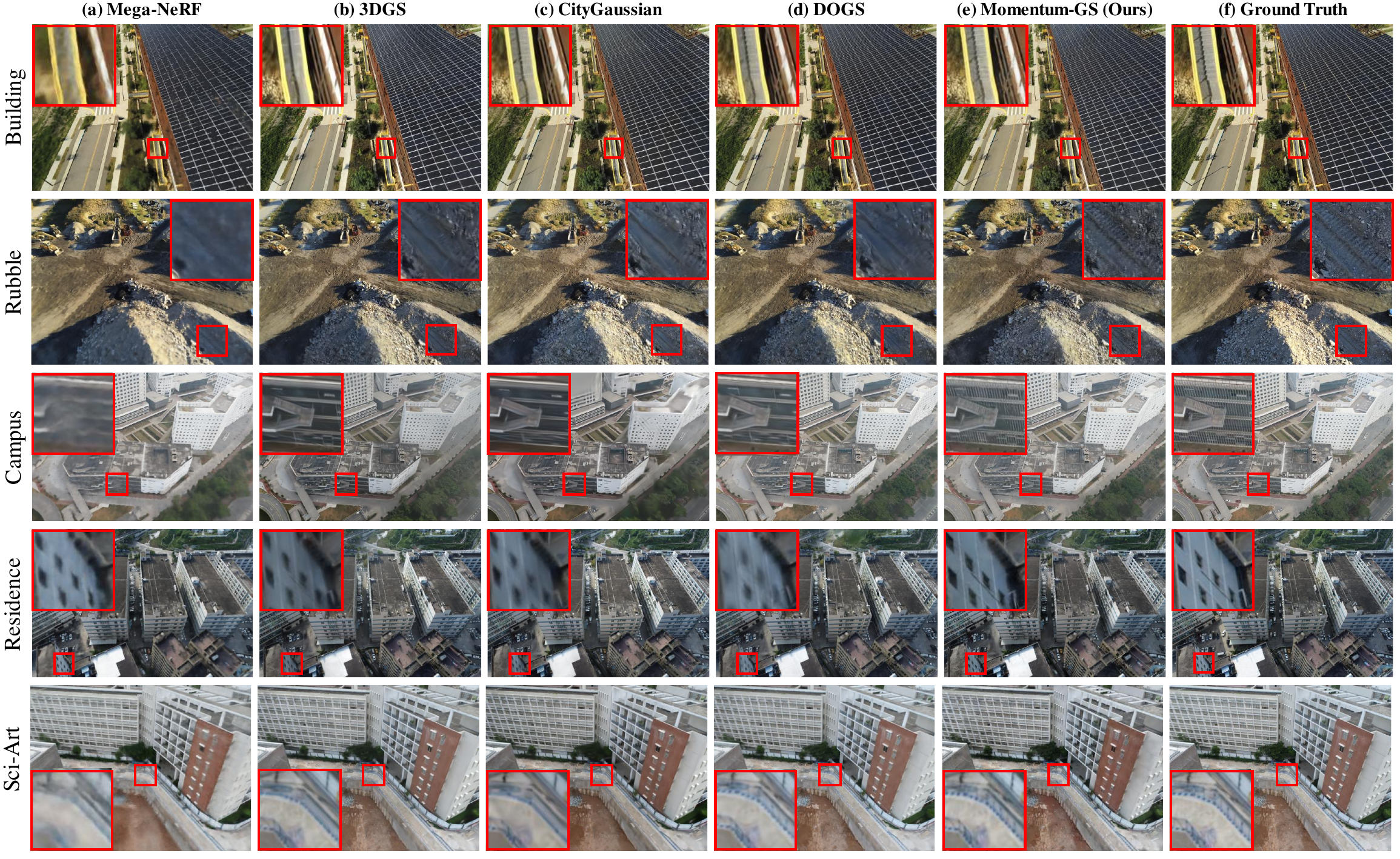}
   \caption{Qualitative comparisons of our \model~with previous methods on five large-scale scenes. \textcolor{red}{Red} insets highlight areas with notable visual differences. Our approach (e) captures finer details and more accurately represents textures, producing visual reconstructions closer to the ground truth (f). In contrast, previous methods exhibit noticeable artifacts, blur, or inconsistencies in these regions.}
   \label{fig:qualitative_comparison}
   \vspace{-10pt}
\end{figure*}

\paragraph{Quantitative Results.}
In Table~\ref{tab:main_table1} and Table~\ref{tab:main_table2}, we present the quantitative evaluation results across six large-scale scenes. Our proposed \model consistently achieves the best overall performance, substantially outperforming other approaches. These results highlight the capability of our \model in preserving fine details and delivering high-quality renderings.
Notably, NeRF-based methods yield higher PSNR scores on the \textit{Sci-Art} dataset. This phenomenon likely arises due to the inherent blurriness present in the \textit{Sci-Art} source images, perhaps due to out-of-focus capture conditions. Since NeRF-based methods typically generate smoother and blurred reconstructions, their outputs naturally align better with these blurred ground-truth images, resulting in elevated PSNR scores. However, when considering SSIM and LPIPS metrics, Gaussian-based methods, typically our \model, significantly outperform NeRF-based approaches in perceptual quality.

\begin{figure*}[ht]
  \centering
   \includegraphics[width=1\linewidth]{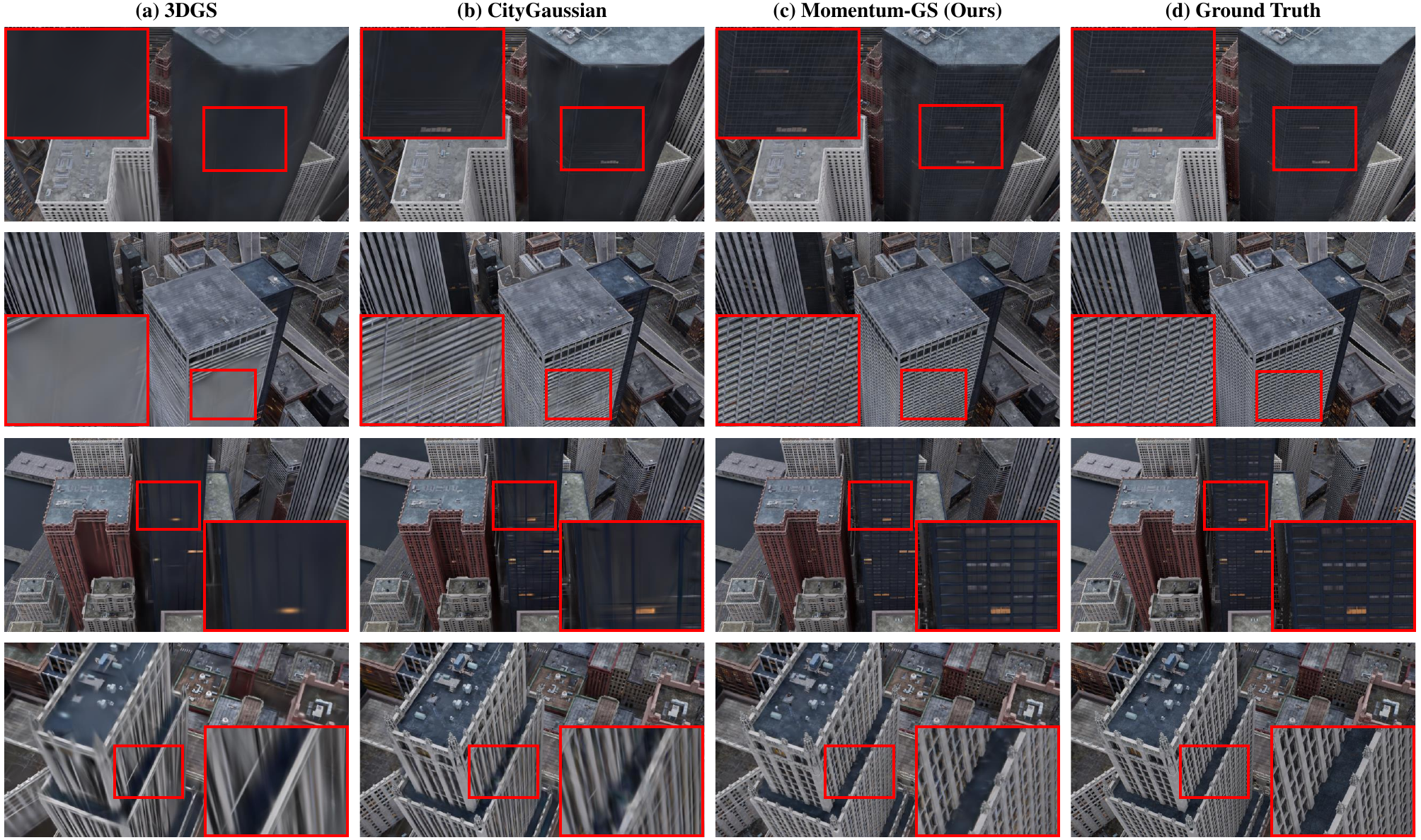}
   \caption{Qualitative comparisons between our \model and other methods on the extremely large-scale urban scene, \textit{MatrixCity}. Our method~(c) demonstrates superior detail preservation in challenging regions, such as building facades and edges, closely resembling the ground truth~(d). In contrast, other approaches exhibit noticeable artifacts, blurring, and loss of structural details in these areas.}

   \label{fig:qualitative_comparison_matrixcity}
   \vspace{-5pt}
\end{figure*}

\vspace{-10pt}
\paragraph{Visualization Results.}
In Figure~\ref{fig:qualitative_comparison} and Figure~\ref{fig:qualitative_comparison_matrixcity}, we provide visual comparisons of reconstruction results across six scenes. Our proposed \model consistently produces sharp and realistic images, demonstrating superior detail preservation and excellent visual clarity across all scenes. In contrast, other methods often suffer from noticeable blurring and structural degradation, especially in complex regions. These qualitative results further highlight the effectiveness of \model in capturing fine-grained details and maintaining overall rendering quality.

\subsection{Ablation Studies}
\paragraph{Parallel training vs. Independent training.} 

In Table~\ref{tab:ablation_training_strategies}, we demonstrate that parallel training (b) achieves better reconstruction quality compared to independent training (c) when the scene is divided into the same number of blocks. This improvement arises from the increased data diversity available to the shared Gaussian decoder. However, direct parallel training is constrained by the requirement that the number of blocks must match the number of available GPUs. Consequently, independent training can further enhance accuracy by utilizing a larger number of blocks, whereas direct parallel training (b) remains limited by GPU availability. As shown in the Table~\ref{tab:ablation_training_strategies}, independent training with eight blocks (d) yields better performance and surpass (b). To overcome this limitation, we introduce scene momentum self-distillation (e), enabling the Gaussian decoder to benefit from increased data diversity while decoupling the number of blocks from the GPU count. Our approach achieves significant accuracy improvements compared to independently training eight blocks. Moreover, incorporating reconstruction-guided block weighting (denoted as "(f) Full") further enhances the overall reconstruction quality.

\begin{table}[ht!]
	\begin{center}
		\resizebox{0.8\linewidth}{!}{
			\begin{tabular}{l|ccc}
				\toprule
				Models &  PSNR \(\uparrow\) & SSIM \(\uparrow\) & LPIPS \(\downarrow\) \\
				\midrule
				w/ PSNR                   & 23.49 & 0.809 & 0.197 \\
				w/ SSIM                   & 23.53 & 0.806 & 0.203 \\
				\midrule
				Full (PSNR + SSIM)      & \textbf{23.65} & \textbf{0.813} & \textbf{0.194} \\
				\bottomrule
			\end{tabular}		
		}
		\caption{Ablation study on different strategy of measuring the reconstruction quality in block weighting.}
		\label{tab:ablation_block_weighting}
	\end{center}
	\centering
    \vspace{-15pt}
\end{table}

\begin{table}[t!]
	\begin{center}
		\resizebox{\linewidth}{!}{
			\begin{tabular}{l|c|ccc}
				\toprule
				Training strategy & \#Block &  PSNR \(\uparrow\) & SSIM \(\uparrow\) & LPIPS \(\downarrow\) \\
				\midrule
				(a) baseline                                     & 1 & 22.25 & 0.742 & 0.272 \\
				(b) w/ Parallel training                         & 4 & 23.10 & 0.790 & 0.221 \\
				(c) w/ Independent training                      & 4 & 22.85 & 0.781 & 0.229 \\
				(d) w/ Independent training                      & 8 & 23.23 & 0.796 & 0.211 \\
				(e) w/ momentum self-distill.                & 8 & 23.56 & 0.806 & 0.205 \\
				\midrule
				(f) Full     & 8 & \textbf{23.65} & \textbf{0.813} & \textbf{0.194} \\
				\bottomrule
			\end{tabular}		
		}
		\caption{Ablation study on different training strategies.}
		\label{tab:ablation_training_strategies}
	\end{center}
	\centering
    \vspace{-20pt}
\end{table}

\vspace{-10pt}
\paragraph{Block weighting.}
In Table~\ref{tab:ablation_block_weighting}, we evaluate different methods for measuring the reconstruction quality of each block. Results show that combining PSNR and SSIM yields higher accuracy than using either alone.

\vspace{-10pt}
\paragraph{Scalability.}
We evaluated our method on various numbers of divided blocks, keeping the GPU count constant at four. As shown in Table~\ref{tab:abation_block_num}, reconstruction quality consistently improves with more blocks, demonstrating our method's scalability under limited GPU resources.

\begin{table}[t!]
	\begin{center}
		\resizebox{\linewidth}{!}{
			\begin{tabular}{l|c|ccc}
				\toprule
				Method & \#Block &  PSNR \(\uparrow\) & SSIM \(\uparrow\) & LPIPS \(\downarrow\) \\
				\midrule
				CityGaussian  & 32   & 28.61 & 0.868 & 0.205 \\
				\model (Ours)  & 4   & 28.93 & 0.870 & 0.203 \\
				\model (Ours) & 8   & 29.11 & 0.881 & 0.180 \\
				\model (Ours) & 16  & 29.15 & 0.884 & 0.172 \\
				\bottomrule
			\end{tabular}		
		}
        \caption{Ablation study on the different number of divided blocks.}
		\label{tab:abation_block_num}
	\end{center}
	\centering
    \vspace{-20pt}
\end{table}

%% file: sec/5_conclusion.tex
\section{Conclusion}
\label{sec:conclusion}
In this paper, we have introduced \model, a novel momentum-based self-distillation framework that notably enhances 3D Gaussian Splatting for large-scale scene reconstruction. The core of \model is a momentum-updated teacher Gaussian decoder, which serves as a stable global reference to guide parallel training blocks, effectively promoting spatial consistency and coherence across the reconstructed scene. We further introduce a reconstruction-guided block weighting mechanism, which dynamically adjusts the emphasis on each block based on reconstruction quality, further improving overall consistency. Our approach leverages hybrid representations, integrating both implicit and explicit features, to enable flexible scaling that decouples the number of blocks from GPU constraints. Experimental results demonstrate the strong capability of hybrid representations and momentum-based self-distillation for robust, large-scale 3D scene reconstruction.

\section*{Acknowledgements} 
This work was supported by Guangdong Natural Science Funds for Distinguished Young Scholar (No. 2025B1515020012) and Shenzhen Science and Technology Program (JCYJ20240813111903006).

%% file: sec/6_suppl.tex
\clearpage
\appendix
\renewcommand{\thesection}{\Alph{section}}
\setcounter{page}{0}
\maketitlesupplementary

\section{More Details}
\textbf{Scene partition.}
(1) Criteria: 
The scene is first equally divided along the x-axis and then along the z-axis, with each block having the same area.
Corresponding views are selected based on visibility. (2) Initialization: Each block is initialized from the same point cloud generated by COLMAP, but only the assigned part and overlapping boundary are reconstructed. 
(3) Views selection: Views at the boundaries are selected based on visibility, 
and each block reconstructs an extended region to ensure better reconstruction quality at the boundary area. 


\noindent\textbf{Motivation of momentum updates.}
The momentum-based update provides stable, global guidance, allowing each block’s Gaussian decoder to effectively leverage the broader scene context, thereby significantly enhancing reconstruction consistency. As demonstrated in Table~\ref{table:momentum_value}, using a momentum value of 0.9 outperforms a setting without momentum updates.

\section{More Ablation Study}

\textbf{Effectiveness of self-distillation.}
As shown in Table~\ref{tab:ablation_training_strategies_supp}, we performed additional experiments to validate the effectiveness of our self-distillation approach: (1) As shown in setting (b), extending parallel training to 8 blocks with 8 GPUs improved the reconstruction quality. (2) Alternating training across blocks every 500 iterations, using 4 GPUs to train 8 blocks in parallel (setting (c)), slightly decreased the reconstruction quality compared with setting (b). (3) Incorporating our momentum-based self-distillation into setting (c) enhanced the reconstruction quality (setting (d)), clearly demonstrating the effectiveness of our proposed method.

\begin{table}[!htp]
	\begin{center}
		\resizebox{\linewidth}{!}{
			\begin{tabular}{l|c|c|ccc}
				\toprule
				Training strategy & \#Block & \#GPU &  PSNR \(\uparrow\) & SSIM \(\uparrow\) & LPIPS \(\downarrow\) \\
				\midrule
				(a) w/ Parallel training                         & 4 & 4 & 23.10 & 0.790 & 0.221 \\
				\underline{(b) w/ Parallel training}                         & 8 & 8 & 23.34 & 0.800 & 0.210 \\
				\underline{(c) w/ Parallel training (alternating)}           & 8 & 4 & 23.17 & 0.797 & 0.211 \\
				(d) w/ momentum self-distill.                    & 8 & 4 & 23.56 & 0.806 & 0.205 \\
				\midrule
				(e) Full     & 8 & 4 & \textbf{23.65} & \textbf{0.813} & \textbf{0.194} \\
				\bottomrule
			\end{tabular}		
		}
		\caption{Ablation study on different training strategies.}
		\label{tab:ablation_training_strategies_supp}
	\end{center}
	\centering
\end{table}

\noindent\textbf{The weight of consistency loss.}
An ablation study is performed to evaluate the impact of the consistency loss weight $\lambda_{consistency}$. As reported in Table~\ref{tab:ablation_consistency}, the results indicate that model performance remains stable across a wide range of $\lambda_{consistency}$ values.

\begin{table}[!htp]
	\begin{center}
		\resizebox{\linewidth}{!}{
			\begin{tabular}{l|ccc|ccc}
				\toprule
				Scene   &   \multicolumn{3}{c|}{\emph{Building}}  &   \multicolumn{3}{c}{\emph{Rubble}} \\
				\midrule
				$\lambda_{consistency}$ &  PSNR \(\uparrow\) & SSIM \(\uparrow\) & LPIPS \(\downarrow\) &  
                PSNR \(\uparrow\) & SSIM \(\uparrow\) & LPIPS \(\downarrow\) \\
				\midrule
				1  & 23.53 & 0.808 & 0.201     & 26.51 & 0.816 & 0.210  \\
				10  & 23.63 & 0.810 & 0.200     & 26.62 & 0.821 & 0.204    \\
				50  & \textbf{23.65} & \textbf{0.813} & \textbf{0.194}     & 26.66 & 0.826 & 0.200   \\
				100  & 23.63 & 0.810 & 0.197     & \textbf{26.69} & \textbf{0.829} & \textbf{0.198}    \\
				\bottomrule
			\end{tabular}		
		}
        \caption{Ablation study on $\lambda_{consistency}$.}
		\label{tab:ablation_consistency}
	\end{center}
	\centering
\end{table}

\begin{table}[!htp]
    \begin{center}
        \resizebox{0.7\linewidth}{!}{
            \begin{tabular}{l|ccc}
                \toprule
                Momentum values & PSNR \(\uparrow\) & SSIM \(\uparrow\) & LPIPS \(\downarrow\) \\ 
                \midrule
                0.0 & 23.44 & 0.806 & 0.203 \\
                0.5 & 23.59 & 0.808 & 0.201 \\ 
                0.7 & 23.62 & 0.810 & 0.198 \\ 
                0.9 (default) & 23.65 & 0.813 & 0.194 \\ 
                0.95 & 23.50 & 0.806 & 0.201 \\ 
                0.99 & 22.06 & 0.741 & 0.254 \\ 
                \bottomrule 
                \end{tabular}
        }
        \caption{Comparison between different momentum values.}
        \label{table:momentum_value} 
    \end{center}
    \centering
\end{table}

\noindent\textbf{Momentum value.} We ablated momentum value $m$ and Table~\ref{table:momentum_value} shows that our model is robust to variations. The reconstruction quality show minimal differences, with the best performance achieved at m=0.9 (our default setting).

\begin{figure*}[htbp]
  \centering
   \includegraphics[width=1\linewidth]{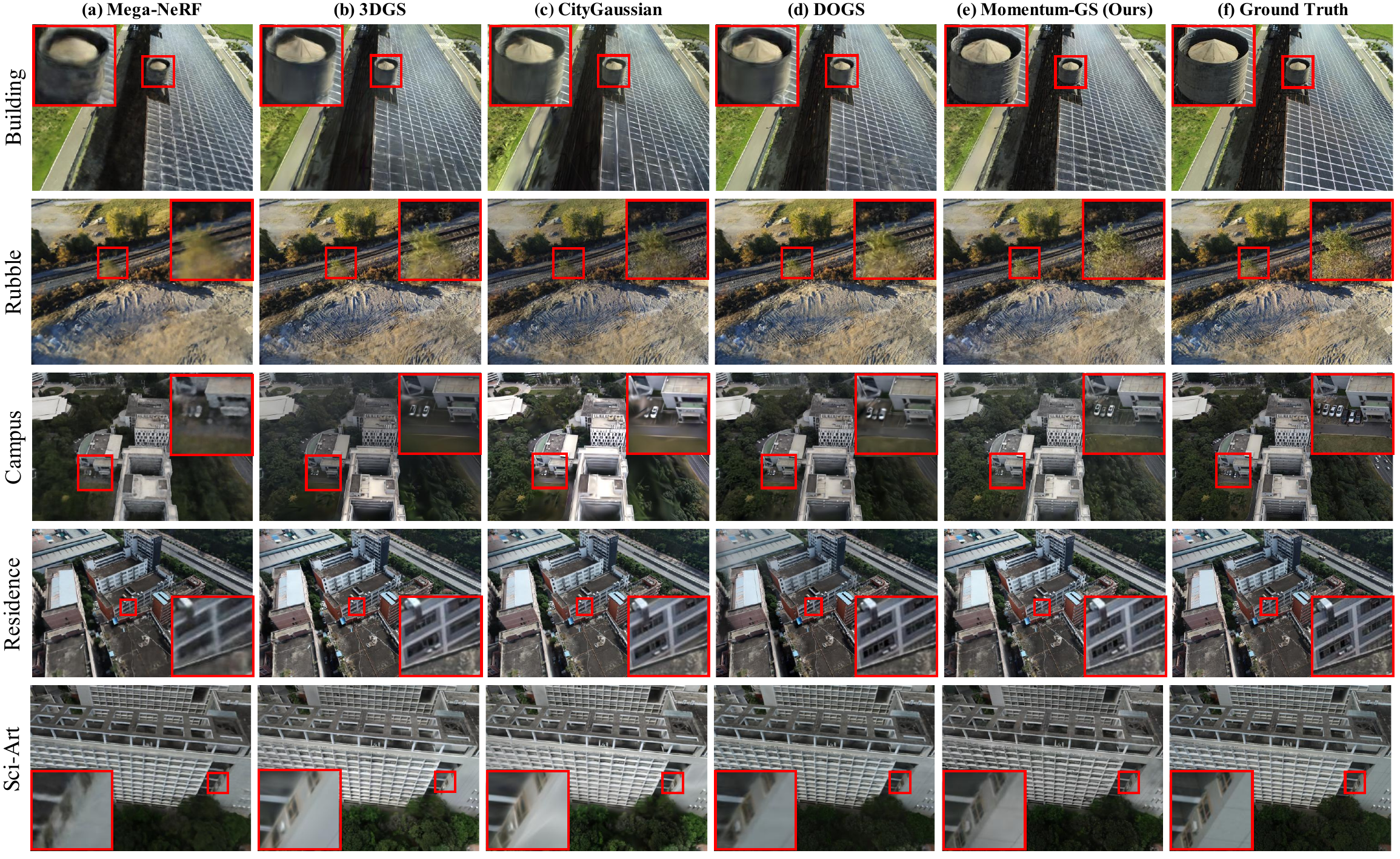}
   \caption{Qualitative comparisons of our \model and prior methods across four large-scale scenes.}
   \label{fig:qualitative_comparison_supp}
\end{figure*}

\section{Quantitative Evaluation}

\textbf{VRAM.}
We report the peak VRAM usage during inference across five large-scale scenes, as shown in Table~\ref{tab:vram}. Despite achieving superior reconstruction quality, our method requires less VRAM compared to the purely 3DGS-based approach. The VRAM usage, measured in MB, highlights the efficiency of our method. Notably, as scene complexity increases (\eg, in MatrixCity), the advantages of our method become even more pronounced. 

\begin{table}[h!]
	\begin{center}
		\resizebox{\linewidth}{!}{
			\begin{tabular}{l|ccccc}
				\toprule
				Scene & Building & Rubble & Residence & Sci-Art & MatrixCity \\
				\midrule
				CityGaussian  & 8977 & 5527 & 6494 & \textbf{2726} & 14677 \\
				\model (Ours) & \textbf{5830} & \textbf{4106} & \textbf{6419} & 6647 & \textbf{4616} \\
				\bottomrule
			\end{tabular}		
		}
        \caption{Peak VRAM usage (in MB) during inference.}
		\label{tab:vram}
	\end{center}
	\centering
\end{table}

\noindent\textbf{Storage.}
We report the storage usage across five large-scale scenes, as shown in Table~\ref{tab:storage}. Leveraging our hybrid representation, our method significantly reduces the number of parameters required for storage compared to purely 3DGS-based methods. This reduction is especially notable in larger and more complex scenes, such as MatrixCity, where the storage savings are most substantial. Notably, as scene complexity increases (\eg, in MatrixCity), the advantages of our method become even more pronounced, demonstrating its effectiveness in handling challenging scenarios. For clarity and consistency, storage usage is reported in GB.
\begin{table}[h!]
	\begin{center}
		\resizebox{\linewidth}{!}{
			\begin{tabular}{l|ccccc}
				\toprule
				Scene & Building & Rubble & Residence & Sci-Art & MatrixCity \\
				\midrule
				CityGaussian  & 3.07 & 2.22 & 2.49 & \textbf{0.88} & 5.40 \\
				\model (Ours) & \textbf{2.45} (20.2\%$\downarrow$) & \textbf{1.50} (32.7\%$\downarrow$) & \textbf{2.00} (19.7\%$\downarrow$) & 0.97 & \textbf{2.08} (61.5\%$\downarrow$) \\
				\bottomrule
			\end{tabular}		
		}
        \caption{Storage usage (in GB).}
		\label{tab:storage}
	\end{center}
	\centering
\end{table}

\noindent\textbf{Number of primitives.}
We report the number of primitives across five large-scale scenes, as shown in Table~\ref{tab:primitives}.

\begin{table}[h!]
	\begin{center}
		\resizebox{\linewidth}{!}{
			\begin{tabular}{l|ccccc}
				\toprule
				Scene & Building & Rubble & Residence & Sci-Art & MatrixCity \\
				\midrule
				Primitives  & 8.33M & 5.09M & 6.79M & 3.30M & 7.08M \\
				\bottomrule
			\end{tabular}		
		}
        \caption{Primitives counts for each scene.}
		\label{tab:primitives}
	\end{center}
	\centering
\end{table}

\noindent\textbf{Comparison of different implementations of VastGaussian.}
We further compare our method with the unofficial implementation of VastGaussian in Table~\ref{tab:vastgs_unofficial}, which demonstrates improved performance over the results reported in DOGS.

\begin{table}[!htp]
	\begin{center}
		\resizebox{\linewidth}{!}{
			\begin{tabular}{l|ccc|ccc}
				\toprule
				Scene   &   \multicolumn{3}{c|}{\emph{Building}}  &   \multicolumn{3}{c}{\emph{Rubble}} \\
				\midrule
				Metrics &  PSNR \(\uparrow\) & SSIM \(\uparrow\) & LPIPS \(\downarrow\) &  
                PSNR \(\uparrow\) & SSIM \(\uparrow\) & LPIPS \(\downarrow\) \\
				\midrule
				$\text{VastGaussian (DOGS version)}$  & 21.80 & 0.728 & 0.225     & 25.20 & 0.742 & 0.264  \\
				$\text{VastGaussian (Unofficial)}$  & 22.49 & 0.742 & 0.208     & 25.64 & 0.760 & 0.202    \\
				\midrule
				\textbf{Momentum-GS~(Ours)}  & \textbf{23.65} & \textbf{0.813} & \textbf{0.194}      & \textbf{26.66} & \textbf{0.826} & \textbf{0.200}    \\
				\bottomrule
			\end{tabular}		
		}
        \caption{Comparison of different implementations of VastGaussian.}
		\label{tab:vastgs_unofficial}
	\end{center}
	\centering
\end{table}

\section{More Visual Comparisons}

We provide additional visual comparisons for the Building, Rubble, Residence, and Sci-Art scenes in Figure~\ref{fig:qualitative_comparison_supp}. Our method consistently reconstructs finer details across these scenes. Notably, our approach demonstrates a superior ability to reconstruct luminance, as illustrated by the Sci-Art example shown in Figure~\ref{fig:qualitative_comparison_supp}. While NeRF-based methods are capable of capturing luminance by leveraging neural networks to learn global features such as lighting, they tend to produce blurrier results compared to 3DGS-based methods. This underscores the effectiveness of our hybrid representation, which combines the strengths of both NeRF-based and 3DGS-based approaches.